\documentclass[10pt,twocolumn,letterpaper]{article}

\usepackage[pagenumbers]{cvpr} 
\usepackage{multirow}

\definecolor{cvprblue}{rgb}{0.21,0.49,0.74}
\usepackage[pagebackref,breaklinks,colorlinks,allcolors=cvprblue]{hyperref}
\usepackage{xcolor}
\usepackage{bbding}

\def\paperID{*****} 
\def\confName{arxiv}
\def\confYear{2026}
\newcommand\dataset{AEGIS}
\title{\dataset: Exploring the Limit of World Knowledge Capabilities for Unified Mulitmodal Models}

\makeatletter
\newcommand\blfootnote[1]{
  \begingroup
  \renewcommand\thefootnote{}\footnote{#1}
  \addtocounter{footnote}{-1}
  \endgroup
}
\makeatother

\author{
    \textbf{Jintao Lin}\textsuperscript{1$*$},
    \textbf{Bowen Dong}\textsuperscript{2$*$},
    \textbf{Weikang Shi}\textsuperscript{3},
    \textbf{Chenyang Lei}\textsuperscript{4},\\
    \textbf{Suiyun Zhang}\textsuperscript{4},
    \textbf{Rui Liu}\textsuperscript{4},
    \textbf{Xihui Liu}\textsuperscript{1\dag}\\
    [2mm] 
    \textsuperscript{1}University of Hong Kong \quad
    \textsuperscript{2}The Hong Kong Polytechnic University \\
    \textsuperscript{3}The Chinese University of Hong Kong \quad
    \textsuperscript{4}Huawei Research\\
    [2mm]
}

\begin{document}

\twocolumn[{%
\renewcommand\twocolumn[1][]{#1}%
\maketitle
\vspace{-2em}
\begin{center}
   \captionsetup{type=figure}
   \includegraphics[width=0.99\textwidth]{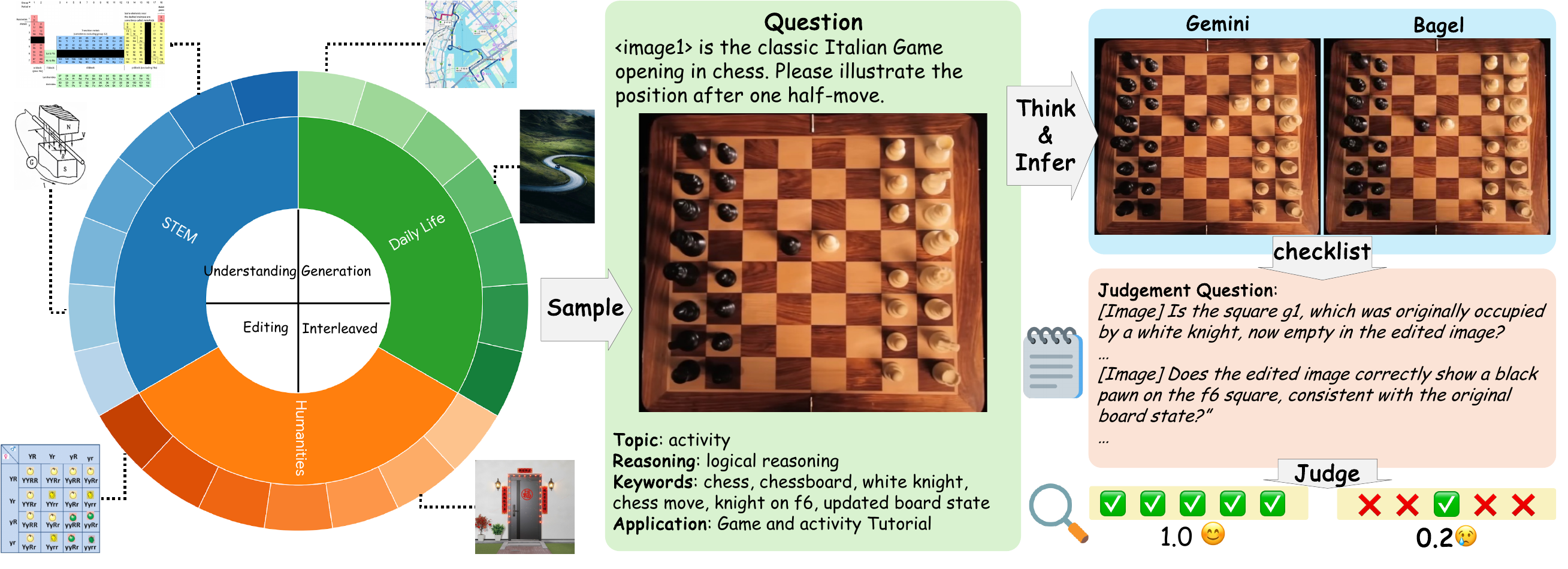}
   \vspace{-1em}
   \captionof{figure}{
  Illustration of {\dataset} benchmark. The contribution of {\dataset} includes 1) a comprehensive and challenging benchmark with four differenct visual understanding and generation tasks, covering a board knowledge aspect (\emph{i.e.}, world knowledge); 2) a deterministic checlist-based evaluation protocol for concrete evaluation results; 3) empirical analysis for state-of-the-art unified multimodal models and other generative models to reveal the vulnerability on world knowledge and reasoning. 
   }\label{fig:teaser}
\end{center}%
}]

\blfootnote{\noindent$^{*}$Equal contribution, in random order. \textsuperscript{\dag}Corresponding author: Xihui Liu \textless xihuiliu@eee.hku.hk\textgreater.}

\begin{abstract}
The capability of Unified Multimodal Models (UMMs) to apply world knowledge across diverse tasks remains a critical, unresolved challenge. Existing benchmarks fall short, offering only siloed, single-task evaluations with limited diagnostic power. To bridge this gap, we propose {\dataset} (\emph{i.e.}, \textbf{A}ssessing \textbf{E}diting, \textbf{G}eneration, \textbf{I}nterpretation-Understanding for \textbf{S}uper-intelligence), a comprehensive multi-task benchmark covering visual understanding, generation, editing, and interleaved generation. {\dataset} comprises 1,050 challenging, manually-annotated questions spanning 21 topics (including STEM, humanities, daily life, etc.) and 6 reasoning types. 
To concretely evaluate the performance of UMMs in world knowledge scope without ambiguous metrics, we further propose Deterministic Checklist-based Evaluation (DCE), a protocol that replaces ambiguous prompt-based scoring with atomic ``Y/N'' judgments, to enhance evaluation reliability. Our extensive experiments reveal that most UMMs exhibit severe world knowledge deficits and that performance degrades significantly with complex reasoning. Additionally, simple plug-in reasoning modules can partially mitigate these vulnerabilities, highlighting a promising direction for future research.
These results highlight the importance of world-knowledge-based reasoning as a critical frontier for UMMs.
\end{abstract}    
\section{Introduction}
\label{sec:intro}
The rapid development of multimodal large language models (MLLMs)~\cite{Qwen25VL,zhu2025internvl3,hurst2024gpt4o,comanici2025gemini,wei2025deepseekocr} and generative models (\emph{e.g.} text-to-image diffusion models)~\cite{labs2025flux,cao2025hunyuanimage,zhao2023unicontrolnet,podell2023sdxl} has achieved remarkable success in visual understanding and generation tasks and applications. The success of these separate models enhances the foundation of artificial intelligence, and inspires researchers to explore a combined framework towards a ``one-for-all'' paradigm~\cite{chen2025janus,hurst2024gpt4o,wang2024emu3}.
Therefore, Unified Multimodal Models (UMMs)~\cite{wu2025qwenimage,wang2024emu3,geng2025xomni,wu2025omnigen2,jiao2025unitoken,deng2025bagel,chen2025blip3o,wang2025ovisu1} have become a significant trend in artificial intelligence. UMMs can simultaneously handle multiple modalities and tasks within a single network, making them highly versatile and compatible with various downstream applications. 
\begin{table*}[ht!]
    \centering
    \vspace{-1em}
    \caption{
    Comparison of {\dataset} with existing world knowledge evaluation benchmarks.
    U/G/E/I in the Tasks column indicate visual understanding, generation, editing, and interleaved generation tasks, respectively. 
    {\dataset} offers superior knowledge and reasoning type coverage, as well as a deterministic and reliable evaluation metric.
    }
\setlength{\tabcolsep}{1pt} %
    \renewcommand{\arraystretch}{3.5}%
    { \fontsize{8.3}{3}\selectfont{
\begin{tabular}{l|cccc|ccc|cccccc|cc}
   \toprule
   \multirow{2}{*}{\bf{Benchmarks}} & \multicolumn{4}{c|}{\bf{Tasks}} & \multicolumn{3}{c|}{\bf{Domains}} & \multicolumn{6}{c|}{\bf{Reasoning Types}} &  \multicolumn{2}{c}{\bf{Evaluation}}  \\ \cline{2-16}
   &U &G&E&I&{STEM} & {Humanity} & {Daily Life} &{Spatial} & {Temporal} & {Casual} & {Comparative} & {Analogical} & {Logical} & {Eval Type} & {Concrete}  \\
   \hline
   WISE~\cite{niu2025wise} & \XSolidBrush & \Checkmark & \XSolidBrush & \XSolidBrush & \Checkmark & \Checkmark & \XSolidBrush & \Checkmark  & \Checkmark & \XSolidBrush & \XSolidBrush  & \XSolidBrush & \Checkmark & Score-based & \XSolidBrush \\
   RISE~\cite{zhao2025rise} & \XSolidBrush & \XSolidBrush & \Checkmark & \XSolidBrush & \Checkmark & \XSolidBrush & \Checkmark & \Checkmark  & \Checkmark & \Checkmark & \XSolidBrush  & \XSolidBrush & \Checkmark & Score-based & \XSolidBrush \\
   KIRS-Bench~\cite{wu2025kris} & \XSolidBrush & \XSolidBrush & \Checkmark & \XSolidBrush & \Checkmark & \XSolidBrush & \Checkmark & \XSolidBrush  & \XSolidBrush & \Checkmark & \XSolidBrush  & \XSolidBrush & \Checkmark & Score-based & \XSolidBrush \\
   T2I-ReasonBench~\cite{sun2025t2i} & \XSolidBrush & \Checkmark & \XSolidBrush & \XSolidBrush & \Checkmark & \XSolidBrush & \Checkmark & \XSolidBrush  & \XSolidBrush & \Checkmark & \XSolidBrush  & \XSolidBrush & \Checkmark & Score-based & \XSolidBrush \\
   R2I-Bench~\cite{chen2025r2i} & \XSolidBrush & \Checkmark & \XSolidBrush & \XSolidBrush & \Checkmark & \XSolidBrush & \Checkmark & \XSolidBrush  & \XSolidBrush & \Checkmark & \Checkmark  & \Checkmark & \XSolidBrush & Score-based & \XSolidBrush \\
   WorldGenBench~\cite{zhang2025worldgenbench} & \XSolidBrush & \Checkmark & \XSolidBrush & \XSolidBrush & \Checkmark & \XSolidBrush & \Checkmark & \XSolidBrush  & \XSolidBrush & \XSolidBrush & \XSolidBrush  & \XSolidBrush & \XSolidBrush & Checklist-based & \XSolidBrush \\
   GIR-Bench~\cite{li2025gir} & \Checkmark & \Checkmark & \Checkmark & \XSolidBrush & \XSolidBrush & \XSolidBrush & \Checkmark & \Checkmark  & \XSolidBrush & \Checkmark & \Checkmark  & \Checkmark & \Checkmark & Score-based & \XSolidBrush \\
    \hline
    \bf{AEGIS (Ours)} & \Checkmark & \Checkmark & \Checkmark & \Checkmark & \Checkmark & \Checkmark & \Checkmark & \Checkmark  & \Checkmark & \Checkmark & \Checkmark  & \Checkmark & \Checkmark & Checklist-based  & \Checkmark  \\
   \bottomrule
\end{tabular}
}
}
\vspace{-1em}
\label{table:dataset_comp}
\end{table*}
However, real-world applications (\emph{e.g.}, AI assistants~\cite{mialon2023gaia,besta2025affordable} and intelligent customer service~\cite{wang2025ecom,fu2020ics}) are fielding increasingly diverse and complex user requests, which often require sophisticated reasoning and extensive world knowledge to answer accurately. 
To meet the demands of these applications, which span a wide array of domains and downstream tasks~\cite{niu2025wise,deng2025bagel}, models must adeptly translate a deep and broad command of world knowledge into high-fidelity text and visual outputs. This capability is critical for providing accurate responses, reducing hallucination, and minimizing the cost of manual rework.


Although numerous benchmarks have been introduced to evaluate MLLMs, generative models, and UMMs, they mostly focus on common tasks (\emph{e.g.}, standard visual question answering)~\cite{mme,seedbench} or simple object and scene generation~\cite{ghosh2023geneval,hu2024dpgbench} and editing~\cite{ye2025imgedit,liu2025gedit}. While recent work has attempted to increase complexity by introducing ``corner knowledge'' topics~\cite{mmiq,guo2025rbench} or reasoning challenges~\cite{sun2025t2i,zhang2025worldgenbench}, they still suffer from three fundamental limitations.
1) Nearly all existing benchmarks are confined to single-task evaluation. As generative models evolve to UMMs, this ``siloed'' assessment fails to measure the critical inter-task performance gaps in UMMs. 
2) They often lack the fine-grained diagnostic capabilities required to localize the source of model failures.
That is, it remains difficult to discern whether poor performance stems from the understanding (LLM) component or the generative module. 
The difficulty and scope of existing benchmarks are often inadequate to comprehensively probe the depth and breadth of a model's world knowledge and complex reasoning abilities.
Therefore, a more comprehensive, challenging, and detailed benchmark, equipped with a novel evaluation protocol, is urgently needed to truly assess the capabilities of modern UMMs.

To address these limitations, as shown in Fig.~\ref{fig:teaser}, we propose {\dataset} (\emph{i.e.}, \textbf{A}ssessing \textbf{E}diting, \textbf{G}eneration, \textbf{I}nterpretation-Understanding for \textbf{S}uper-intelligence), a comprehensive and challenging multi-task benchmark for unified multimodal models and related generative models. Specifically, {\dataset} comprises 1,050 manually annotated and verified questions spanning 21 detailed topics in STEM, the humanities, and daily life. Each question is paired with a corresponding reference answer and keywords, enabling in-depth analysis. Through a human-in-the-loop data construction, refinement, and annotation procedure, {\dataset} covers six distinct reasoning types and assesses four different tasks: visual understanding, generation, editing, and interleaved generation.

Based on the abundant questions, we aim to automatically and accurately evaluate models. Nevertheless, existing scoring-based 'LLM-as-a-Judge' methods~\cite{niu2025wise,sun2025t2i} face two fundamental limitations. The first is heuristic and limited scoring metrics, and the second is ambiguous LLM scores. Although a series of general scoring metrics (\emph{e.g.}, realism and image quality) are widely used in existing benchmarks, they still appear vague, coarse-grained, and lacking in explanatory power. To tackle these issues, we propose a deterministic checklist-based evaluation (DCE).
DCE uses an MLLM (\emph{e.g.}, GPT-4o~\cite{hurst2024gpt4o}) to process a reference response and its corresponding keywords, generating a series of atomic ``Y/N'' judgment questions. Each question is strictly tied to a deterministic key part of the reference. This approach simplifies the complex task of judgment into verifiable steps, thereby improving the reliability of the LLM-as-a-Judge paradigm. AEGIS then effectively measures a UMM's performance by calculating the average percentage of ``yes'' judgments its response receives.

Enabled by our novel data annotation and evaluation protocols, as shown in Table~\ref{table:dataset_comp}, {\dataset} provides significant diagnostic utility compared with existing benchmarks. Firstly, {\dataset} significantly facilitates cross-task evaluation, analyzing correlations between understanding, generation, and editing, rather than assessing tasks in silos. Furthermore, {\dataset} enables in-depth diagnostics by probing different types of reasoning to reveal vulnerabilities and localize component-level deficits. Finally, {\dataset} grounds its assessment in real-world applications, providing a dual (\emph{i.e.}, academic and practical) analysis of model robustness to enhance deployment readiness.


Our extensive evaluation on {\dataset}, covering a wide range of open-source and closed-source models, systematically identifies their respective strengths and weaknesses.
Specifically, most UMMs, with the notable exception of Gemini Nano Banana~\cite{comanici2025gemini}, exhibit severe deficits in world knowledge. Furthermore, performance degrades considerably across all models when complex reasoning is introduced. On a positive note, we demonstrate that integrating simple plug-in reasoning modules can partially mitigate these deficits, suggesting a promising direction for future UMM development.
%
In conclusion, the contributions of this paper are as follows:
\begin{itemize}
    \item We propose \dataset, the first comprehensive and challenging benchmark to simultaneously assess visual understanding, generation, editing, and interleaved generation tasks, covering an extremely broad spectrum of world knowledge.
    \item We propose a deterministic checklist-based evaluation method that uses a series of yes-or-no questions as constraints to assess the correctness of a generated response. This approach provides more reliable judgments than existing methods.
    \item Extensive experimental results on both UMMs and generative models reveal a common vulnerability in their understanding and generation of world knowledge. These findings offer critical insights for the development of future models.
\end{itemize}
\section{Related Work}
\label{sec:related_work}
\subsection{Unified Multimodal Models}\label{sec:related_umm}

Recent Unified Multimodal Models (UMMs) have evolved from MLLMs by integrating generation capabilities, typically through shared backbones and external decoders. Architecturally, these UMMs fall into three main categories: diffusion-based models~\cite{yang2025mmada,wang2025fudoki,shi2025muddit,swerdlow2025unified,li2025dual}, purely auto-regressive approaches~\cite{wu2025qwenimage,wang2024emu3,geng2025xomni,wu2025omnigen2,jiao2025unitoken,chen2025blip3o} that use transformers for feature aggregation, and hybrid models~\cite{deng2025bagel,liao2025mogao,ma2025janusflow,shi2024lmfusion} that fuse auto-regressive text generation with multi-step image denoising. While these models leverage multi-phase training frameworks, two critical questions remain: first, how understanding capabilities truly bolster generation within these unified systems, and second, the true extent of their capacity for world-knowledge understanding~\cite{guo2025rbench,lin2025generative,dong2025mirage,han2023coremm} and generation~\cite{niu2025wise,sun2025t2i,wu2025kris,chen2025t2vworldbench}. Our paper aims to critically investigate these questions.

\begin{figure*}[t]
\begin{center}
\includegraphics[width=0.9\textwidth]{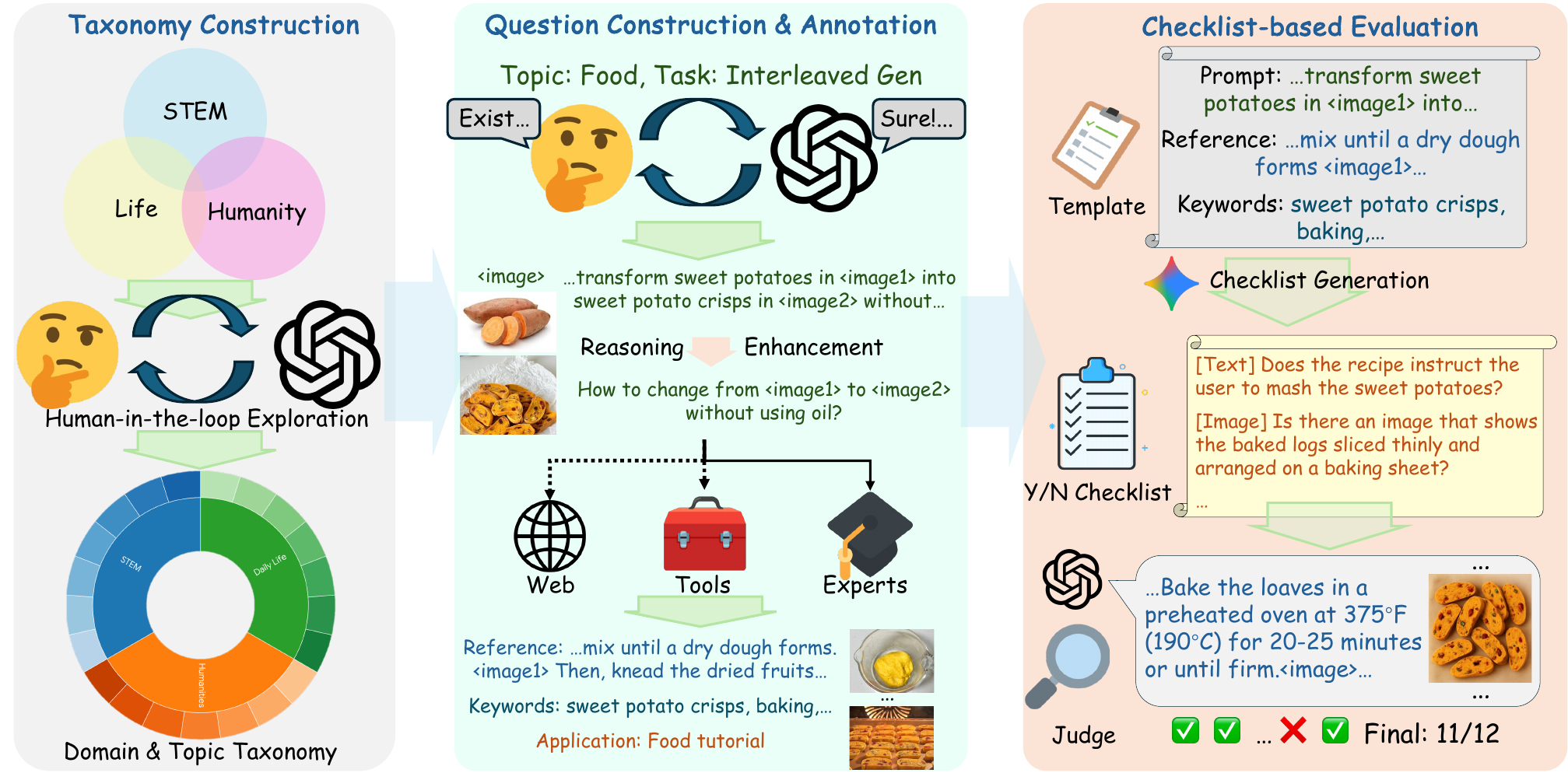}
\end{center}
\vspace{-2em}
\caption{
    Data construction and evaluation pipeline of our proposed {\dataset}. 
    Based on the board taxonomy aspect from human-in-the-loop exploration, {\dataset} features a high-quality data construction procedure, using human-in-the-loop exploration and optimal annotation methods (web, tool, or expert) to create reasoning-enhanced questions. Another key highlight is the novel deterministic checklist-based evaluation (DCE), where an MLLM first generates a checklist of atomic ``Y/N'' questions from the reference answer. A judge MLLM then uses this checklist to produce clear, concrete, and reliable judgments of the model's prediction.
}
\vspace{-1em}
\label{fig:pipeline}
\end{figure*}

\subsection{World Knowledge Benchmarks}\label{sec:related_world_knowledge}
Given the remarkable progress of state-of-the-art generative models in handling common visual understanding~\cite{fu2024mme,mme,seedbench}, generation~\cite{ghosh2023geneval,hu2024dpgbench}, and editing tasks~\cite{liu2025gedit,ye2025imgedit}, recent efforts have begun to assess their capabilities across broader knowledge scopes, \emph{i.e.}, world knowledge~\cite{niu2025wise,fu2024commonsense,hurst2024gpt4o,deng2025bagel}.
To increase the difficulty of world knowledge benchmarks, some works chose to involve specialized or less commonly addressed topics~\cite{puzzlevqa,mmiq,wu2025kris,niu2025wise}. 
For instance, MM-IQ~\cite{mmiq} utilizes graphical IQ test questions to probe the limitations of MLLMs in visual understanding, while WISE~\cite{niu2025wise} introduces instructions from culture and natural science to explore intelligent image generation capabilities.
The second group, which accounts for the majority of recent enhanced world-knowledge benchmarks~\cite{sun2025t2i,zhang2025worldgenbench,li2025gir,dong2025mirage,jiang2025mmecot,guo2025rbench,zhou2025opening,deng2025bagel,wang2025complexbench,huang2024smartedit}, involves increasing the reasoning difficulty of the questions. These benchmarks introduce complex reasoning logics to augment the original instructions, thereby making it more challenging for generative models to correctly comprehend the intended meaning.
Despite their success, existing benchmarks commonly exhibit three critical shortcomings: they tend to over-represent common questions with limited reasoning types while neglecting rare or challenging ones; they might not easy to analyze the influence different tasks may have on one another; and their LLM-based evaluation may include ambiguous metrics (\emph{e.g.}, predicting scores for realism, consistency, and quality metrics). In contrast, our work aims to overcome these specific limitations by various and challenging questions with our checklist-based evaluation protocol.

\section{{\dataset} Benchmark}
\subsection{Dataset Overview}
\label{sec:dataset_overview}
To comprehensively evaluate UMMs and other generative models on visual understanding, generation, editing, and interleaved generation tasks across a \textbf{broad} world knowledge spectrum and \textbf{complex reasoning}, we propose {\dataset}. As detailed in Table~\ref{table:dataset_comp}, {\dataset} covers three general domains (\emph{i.e.}, STEM, humanities, and daily life) with 21 diverse topics. The data statistics are shown in Table~\ref{table:data_distribution}. Each topic contains 15 prompts for visual understanding, generation, and editing, as well as 5 visual interleaved generation questions to measure complex generative capabilities. Furthermore, {\dataset} incorporates six distinct reasoning types into the majority of its prompts, requiring UMMs to possess inherent reasoning capabilities to complete each request. The following sections detail the construction, annotation, and evaluation protocols for {\dataset}.

\begin{table}[htbp]
\centering
\caption{Data distribution of the AEGIS dataset across different topics and tasks, where ``U'' means visual understanding questions, ``G'' means visual generation questions, ``E'' means visual editing questions, and ``I'' means visual interleaved generation questions.}
\label{table:data_distribution}
\vspace{-1em}
\setlength{\tabcolsep}{3pt} %
    \renewcommand{\arraystretch}{2.5}%
    { \fontsize{8.3}{3}\selectfont{
\begin{tabular}{llccccc}
\toprule
\textbf{Domain} & \textbf{Topic} & \textbf{U} & \textbf{G} & \textbf{E} & \textbf{I} & \textbf{Total} \\
\midrule
\multirow{7}{*}{\textbf{STEM}} 
 & Biology & 15 & 15 & 15 & 5 & 50 \\
 & Chemistry & 15 & 15 & 15 & 5 & 50 \\
 & Mathematics & 15 & 15 & 15 & 5 & 50 \\
 & Medicine & 15 & 15 & 15 & 5 & 50\\
 & Physics & 15 & 15 & 15 & 5 & 50\\
 & Astronomy \& Geography & 15 & 15 & 15 & 5 & 50\\
 & IT & 15 & 15 & 15 & 5 & 50\\
\midrule
\multirow{7}{*}{\textbf{Humanities}} 
 & Agriculture & 15 & 15 & 15 & 5 & 50\\
 & History & 15 & 15 & 15 & 5 & 50\\
 & Movie & 15 & 15 & 15 & 5 & 50\\
 & Music & 15 & 15 & 15 & 5 & 50\\
 & Art & 15 & 15 & 15 & 5 & 50\\
 & Culture & 15 & 15 & 15 & 5 & 50\\
 & Architecture & 15 & 15 & 15 & 5 & 50\\
\midrule
\multirow{7}{*}{\textbf{Daily Life}} 
 & Activity & 15 & 15 & 15 & 5 & 50\\
 & Anime & 15 & 15 & 15 & 5 & 50\\
 & Game & 15 & 15 & 15 & 5 & 50\\
 & Photography & 15 & 15 & 15 & 5 & 50\\
 & Engineering & 15 & 15 & 15 & 5 & 50\\
 & Food & 15 & 15 & 15 & 5 & 50\\
 & Traffic & 15 & 15 & 15 & 5 & 50\\
\midrule
\textbf{Total} & \textbf{21 Sub-categories} & \textbf{315} & \textbf{315} & \textbf{315} & \textbf{105} & \textbf{1050}\\
\bottomrule
\end{tabular}
}}
\vspace{-2em}
\end{table}

\subsection{Data Construction and Annotation}
\label{sec:data_construction}
Constructing a benchmark to cover a broad knowledge spectrum presents significant challenges. Our data construction pipeline involved several sequential stages to ensure high quality and comprehensive coverage.

First, we established the foundational taxonomies. For the topic taxonomy, we utilized an LLM~\cite{hurst2024gpt4o} in a human-in-the-loop procedure to progressively explore and define 21 distinct topics. These topics cover common scenarios in STEM, humanities, and daily life, providing a foundation for broad knowledge evaluation. For the reasoning taxonomy, inspired by recent work~\cite{deng2025bagel,wu2025kris,sun2025t2i}, we formulated six reasoning types: spatial reasoning (analyzing location), temporal reasoning (analyzing changes over time), {causal reasoning} (understanding strong causal relations), comparative reasoning (identifying differences), analogical reasoning (identifying similarities), and logical reasoning (analyzing structured relationships).

\begin{figure*}
\begin{center}
\includegraphics[width=0.99\textwidth]{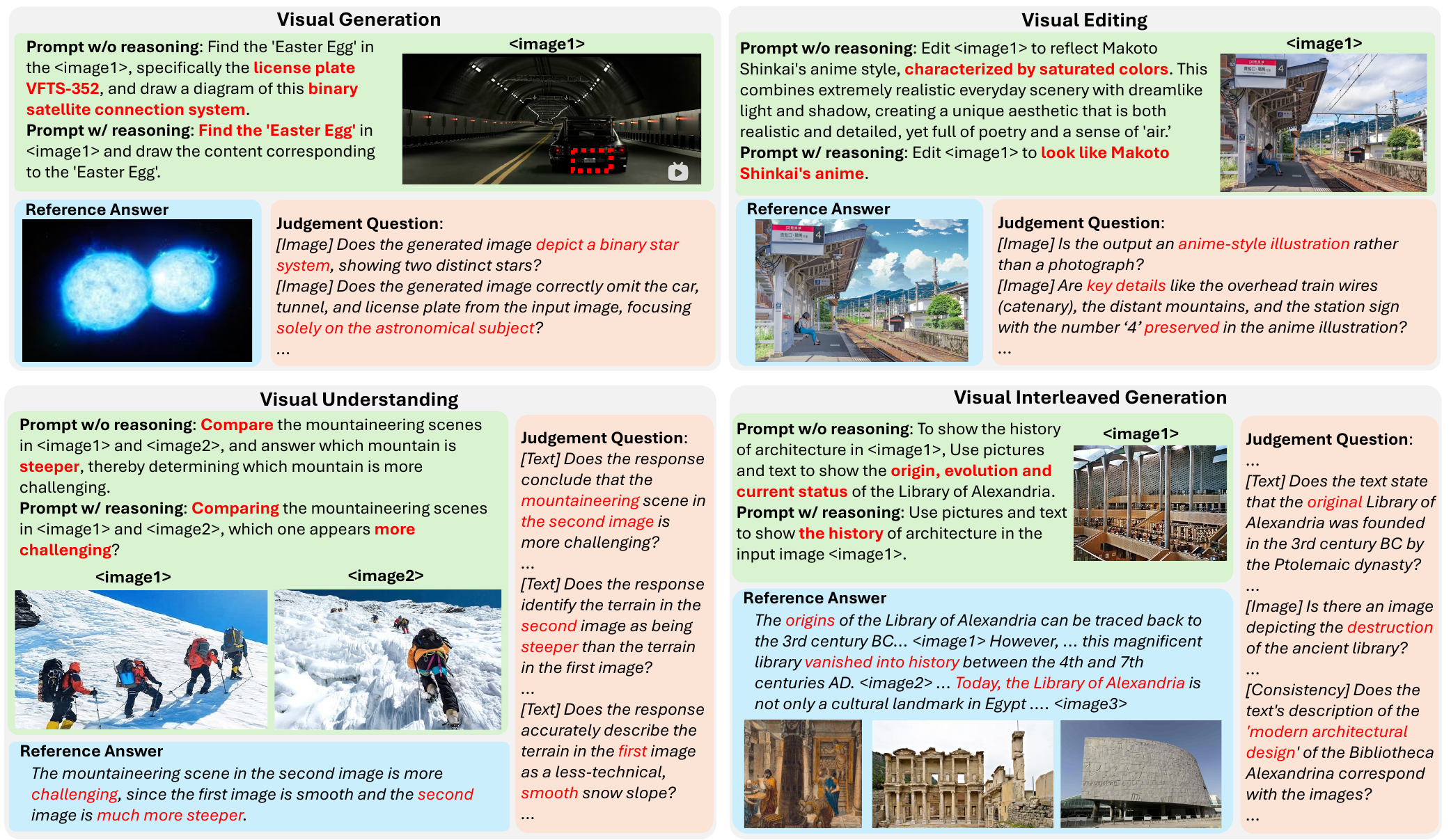}
\end{center}
\vspace{-1.5em}
\caption{
Examples of {\dataset} Benchmark, where red color indicates the key points examined in the question. {\dataset} covers four visual generative tasks with 1,050 reasoning-enhanced questions from 21 different topics, which is useful to explore the generation capabilities of UMMs and other generative models under both broad knowledge aspects (\emph{i.e.}, world knowledge) and different reasoning types. 
}
\vspace{-1.5em}
\label{fig:data_example}
\end{figure*}

Based on these taxonomies, we proceeded to prompt generation using a human-in-the-loop method. With a specific topic and randomly choiced task, we first prompted an LLM to generate a 'clear prompt' for this topic, designed to be unambiguous and free of complex reasoning. If the output was redundant, we provided new 'thinking directions' to guide the LLM until a unique question was constructed. Following this, we manually performed reasoning augmentation by selecting an optimal reasoning type from our taxonomy and converting the 'clear prompt' into a 'reasoning-enhanced' prompt. This resulting prompt mimics the ambiguity and complexity of practical applications, enhancing the real-world utility of {\dataset}.

With the prompts defined, we constructed the necessary inputs and ground-truth annotations based on the choiced task. For prompts requiring image inputs, we built a high-quality dataset via three methods: (1) crawling copyright-free content from the internet, (2) manually creating images using generative models and image editing tools, and (3) commissioning experts to construct the images. This multi-pronged approach ensures high visual fidelity. For ground-truth answers, we used the same three methods to construct reference images for generation and editing tasks. For text-based tasks, an LLM generated a draft response, which was then verified and revised by a human expert. Finally, to facilitate evaluation, we used an MLLM to extract keywords from the clear prompt and its reference answer. These keywords were manually checked to ensure they covered all key points required to solve the question.

\subsection{Deterministic Checklist-based Evaluation}
\label{sec:dce}
Based on the fine-grained annotations, we need a protocol for concrete and reliable evaluation. Existing scoring-based 'LLM-as-a-Judge' methods~\cite{niu2025wise,sun2025t2i,zhang2025worldgenbench} face two fundamental limitations: (1) heuristic and limited scoring metrics, and (2) ambiguous, coarse-grained scores from LLMs. Inspired by the deterministic, point-based judging of competitions (\emph{e.g.}, the International Mathematical Olympiad), where credit is awarded for achieving specific, verifiable steps, we propose the {Deterministic Checklist-based Evaluation (DCE)}. This protocol involves two main phases: judgment question generation and model response judgment.

The generation of the judgment checklist serves as the foundational phase. Given a question (including the clear prompt, optional input image, and reference response) and its keywords, we leverage a state-of-the-art MLLM (\emph{e.g.}, Gemini~\cite{comanici2025gemini}) to extract a series of atomic judgment questions. As shown in Fig.~\ref{fig:pipeline} (right), each question is answerable with only 'yes' or 'no' and focuses on a single, specific aspect of the answer (\emph{e.g.}, a small but detectable modification). This approach reduces ambiguity and judgment difficulty compared to direct scoring. To further enhance quality, we perform manual filtration to remove duplicated or redundant questions ($\sim$20\% of the original set), ensuring the final checklist precisely measures the key points.

{Once this offline-generated checklist is finalized,} it is {employed} for the {model response judgment} phase. We integrate the clear prompt, the model's response, and the checklist into a unified input instruction via a template (detailed in the suppl.). We then leverage an MLLM~\cite{comanici2025gemini} to generate a 'yes' or 'no' answer with a corresponding explanation for each checklist item. Finally, {DCE} effectively measures a UMM's performance by calculating the average percentage of 'yes' judgments its response receives across all questions.

\begin{table*}[t]
\centering
\caption{
    Comprehensive Performance Comparison Across Tasks and Domains.
    Scores are reported for four main tasks, broken down by three domains:
    STEM, Humanity, and Daily Life. ``*'' means models which do not enable interleaved multi-image inputs, and ``\#'' means models which have unsupported tasks (noted by ``-''). Nano Banana and GPT-4o+GPT-Image-1 perform better than others on all tasks. 
}
\label{tab:model_comparison}
\vspace{-0.5em}
\setlength{\tabcolsep}{3pt} %
    \renewcommand{\arraystretch}{3}%
    { \fontsize{8.3}{3}\selectfont{
\begin{tabular}{l| ccc| ccc| ccc |ccc| @{\hspace{1.5em}} c}
\toprule
\multirow{2}{*}{\textbf{Model}} & \multicolumn{3}{c}{\textbf{Understanding}} & \multicolumn{3}{c}{\textbf{Generation}} & \multicolumn{3}{c}{\textbf{Editing}} & \multicolumn{3}{c}{\textbf{Interleaved Generation}} & \multirow{2}{*}{\textbf{Overall}} \\
\cmidrule(lr){2-4} \cmidrule(lr){5-7} \cmidrule(lr){8-10} \cmidrule(lr){11-13}
 & STEM & Humanity & Life & STEM & Humanity & Life & STEM & Humanity & Life & STEM & Humanity & Life & \\
\midrule

\multicolumn{14}{c}{\textbf{Unified Multimodal Models}} \\
\midrule
Gemini Nano Banana & 64.5 & 65.7 & 55.0 & 42.6 & 49.5 & 45.5 & 44.4 & 62.4 & 54.2 & 50.2 & 41.6 & 43.4 & 52.9 \\
GPT-4o+GPT-Image-1 & 52.9 & 50.9 & 46.9 & 38.2 & 51.6 & 42.8 & 39.4 & 53.2 & 45.2 & 38.9 & 34.7 & 33.0 & 45.7 \\
Bagel-7B w/o CoT & 25.5 & 26.9 & 19.3 & 12.1 & 20.6 & 15.3 & 15.0 & 17.6 & 21.2 & 13.0 & 11.9 & 8.2 & 18.5 \\
Bagel-7B w. CoT & 31.8 & 31.7 & 22.0 & 14.9 & 31.2 & 21.3 & 11.6 & 23.5 & 23.1 & 11.8 & 11.9 & 9.9 & 22.3 \\
Ovis-U1$^{*}$ & 26.3 & 31.0 & 17.1 & 12.7 & 25.0 & 16.3 & 19.3 & 27.2 & 25.9 & 12.2 & 12.5 & 8.4 & 21.2 \\
BLIP3o$^{*}$ & 30.8 & 43.3 & 21.7 & 3.7 & 6.3 & 2.7 & 2.6 & 4.4 & 4.5 & 9.4 & 8.3 & 4.0 & 13.7 \\
Qwen-Image & 31.4 & 41.2 & 22.7 & 17.9 & 31.9 & 25.4 & 20.7 & 35.4 & 33.4 & 22.0 & 18.7 & 17.9 & 28.0 \\
Janus-Pro 7B$^{\#}$ & 7.9 & 18.0 & 9.7 & 13.7 & 17.0 & 18.2 & - & - & - & 2.2 & 6.5 & 2.7 & - \\%
Show-o2$^{\#}$ & 15.4 & 26.6 & 11.1 & 16.7 & 24.7 & 22.5 & - & - & - & 5.3 & 8.7 & 3.4 & - \\
Emu-3$^{\#}$ & 3.1 & 8.0 & 2.0 & 8.9 & 19.7 & 14.5 & - & - & - & - & - & - & - \\
\midrule

\multicolumn{14}{c}{\textbf{Understanding MLLMs}} \\
\midrule
Qwen-3-VL 8B & 42.6 & 48.1 & 34.0 & - & - & - & - & - & - & - & - & - & - \\
Kimi-VL-A3B & 30.6 & 36.4 & 23.5 &  - & - & - & - & - & - & - & - & - & - \\
GPT-5 & 67.0 & 57.4 & 60.6 &  - & - & - & - & - & - & - & - & - & - \\
Gemini-2.5-Pro & 72.1 & 77.3 & 63.3 & - & - & - & - & - & - & - & - & - & - \\
\midrule

\multicolumn{14}{c}{\textbf{Image Generation or Editing Models}} \\
\midrule
FLUX.1-Dev$^{*}$ & - & - &-  & 15.4 & 29.2 & 16.8 & - & -& - & - & -& - & -  \\
Step1X-Edit$^{*}$ & - & - & - & - & - & - & 19.8 & 31.9 & 37.1 & - & - & - & - \\
Instruct-Pix2Pix$^{*}$ & - & - & - & - & - & - & 17.3 & 17.6 & 23.5 & - & - & - & - \\
Seedream$^{*}$ & - & - & - & 33.9 & 43.7 & 38.6 & 32.8 & 53.0 & 43.0 & - & - & - & - \\
\bottomrule
\end{tabular}
}}
\vspace{-1.5em}
\end{table*}

\subsection{Joint-Utility Merits}
Enabled by our proposed data annotation and evaluation protocols, {\dataset} demonstrates significant multi-faceted utility and surpasses other benchmarks through the following merits.
First is Inter-task Evaluation. {\dataset} not only allows for a comprehensive evaluation of the four distinct tasks across a broad world knowledge scope but also facilitates a detailed analysis of the correlations between the understanding task and the others. 
Second is the Investigation of Reasoning-Type Effects. By comparing model responses to standard questions against their counterparts involving different reasoning types, researchers can reveal vulnerabilities in the inherent reasoning capabilities of UMMs and generative models. This analysis also facilitates an investigation into which components within the UMMs contribute most to these deficits.
Finally is a Real-World Application Assessment. By linking each data sample to a specific, practical application, {\dataset} facilitates a dual analysis of model vulnerabilities from both an academic and a real-world perspective. This insight is crucial for enhancing model robustness and readiness for deployment.

\section{Experiments}\label{sec:exp}
\subsection{Experimental Setup}
During inference, for black-box models, we directly call corresponding APIs for the final results. 
For open-sourced models, we leverage 4 GPUs to predict the results via transformers~\cite{wolf-etal-2020-transformers} and diffusers~\cite{von-platen-etal-2022-diffusers} toolkit. 
During evaluation, we leverage Gemini-2.5-Pro~\cite{comanici2025gemini} as the default judge model. This model has sufficient fine-grained visual understanding capabilities, which are compatible with the requirements of DCE. 
We will also analyze the choice of judge models in the suppl. to verify our solution. 
\subsection{Baseline Models}\label{sec:baseline_models}
\paragraph{Unified Multimodal Models. }Our primary concern is the capabilities of UMMs. Therefore, during experiments, we introduce a series of black-box or open-sourced UMMs~\cite{deng2025bagel,chen2025blip3o,hurst2024gpt4o,comanici2025gemini}, including Gemini Nano Banana~\cite{comanici2025gemini}, GPT-Image-1~\cite{hurst2024gpt4o}, Bagel with and without CoT~\cite{deng2025bagel}, Ovis-U1~\cite{wang2025ovisu1}, BLIP3o~\cite{chen2025blip3o}, Qwen-Image~\cite{wu2025qwenimage}, Janus-Pro~\cite{chen2025janus}, Show-o2~\cite{xie2025showo2}, and Emu-3~\cite{wang2024emu3}.
\paragraph{Other Generative Models. }In addition to UMMs, for each understanding, generation, and editing task, we also evalute corresponding single-task generative models to assess their capabilities. The evaluated model include multimodal MLLMs (\emph{e.g.}, Qwen3-VL-8B~\cite{Qwen25VL}, GPT-5~\cite{hurst2024gpt4o}, Kii-VL-A3B-Instruct~\cite{team2025kimivl}, and Gemini-2.5-Pro~\cite{comanici2025gemini}) and image generation or editing models~\cite{labs2025flux,liu2025step1xedit,brooks2023instructpix2pix,seedream4}.

\begin{figure*}
\begin{center}
\includegraphics[width=0.99\textwidth]{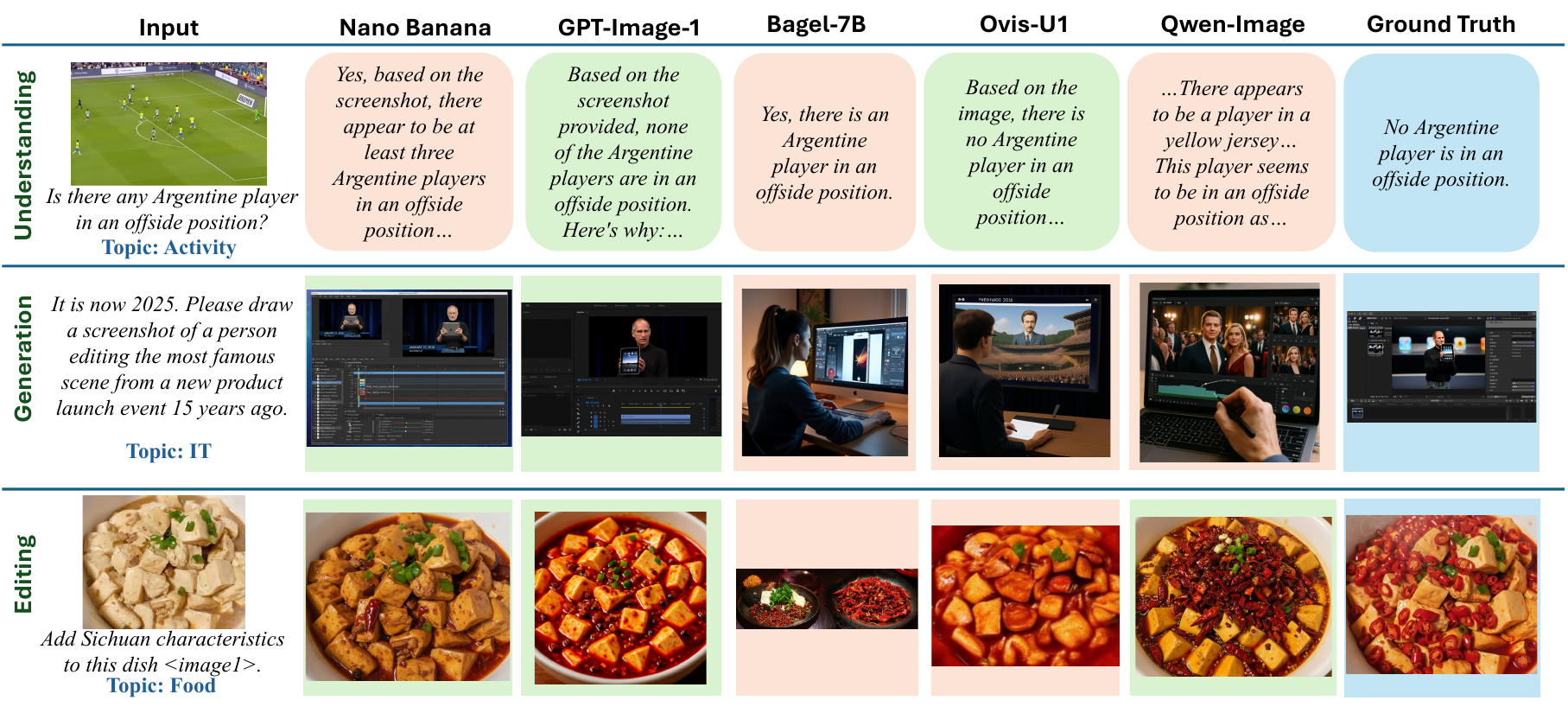}
\end{center}
\vspace{-1em}
\caption{
Visualization of five state-of-the-art UMMs 
on Understanding, Generation, and Editing tasks. Green color means correct responses, and red color means wrong answers. Nana Banana expresses promising image generation and editing quality, and has better main object consistency in editing tasks than others. Meanwhile, open-sourced models don't perform well in these tasks. 
}
\vspace{-1.5em}
\label{fig:visualization_uge}
\end{figure*}

\subsection{Experimental Results}
We conduct experiments among models mentioned in Sec.~\ref{sec:baseline_models}, and illustrate the experimental results in Table~\ref{tab:model_comparison}. These results reveal a stark performance hierarchy. 

\noindent\textbf{General comparison among UMMs. }The leading closed-source UMMs, Gemini Nano Banana and GPT-4o+GPT-Image-1, significantly outperform all other models on all tasks in all domains. Gemini Nano Banana, in particular, establishes the state-of-the-art, demonstrating uniquely strong and balanced capabilities across all four tasks, with standout performance in complex STEM understanding (64.5) and generation (42.6). In sharp contrast, the performance of current open-source UMMs is not promising. Models like Qwen-Image and Bagel-7B, while representing significant open efforts, lag substantially behind their closed-source counterparts. We hypothesize this gap stems from two primary factors: (1) the limited parameter scale of these models, which restricts their capacity to store the extensive world knowledge our benchmark demands, and (2) potential deficiencies in training data quality, especially in lacking the multi-image co-relational data necessary to build robust reasoning capabilities.

\noindent\textbf{Reasoning is useful to refine results}. Compared to Bagel w/ and w/o reasoning, the overall performance improves from 18.5 to 22.3, which indicates that even with a moderate-scale open-sourced model, the understanding and generation capabilities in world knowledge aspects can further enhanced by an external reasoning module. 

\noindent\textbf{More and higher quality training data benefits the world knowledge. }Meanwhile, among understanding MLLMs, Gemini-2.5-Pro and GPT-5 achieve 70.9 and 61.7 on understanding tasks, which is consistent with corresponding UMMs results. Surprisingly, the open-sourced Qwen-3-VL with only 8B parameters also achieves 42.6 on STEM knowledge understanding and 48.1 on Humanity knowledge understanding. In contrast, UMMs which leverage Qwen2.5-VL-7B as a backbone (\emph{e.g.}, Qwen-Image~\cite{wu2025qwenimage} and BLIP3o~\cite{chen2025blip3o}) have much lower performance on understanding tasks. Since the model parameter number and macro design of these models are nearly the same, one can conclude that better (\emph{i.e.}, more fine-grained, abundant, and detailed) multimodal pretraining data benefits world knowledge understanding performance. 

\noindent\textbf{Understanding capability restricts the upper bound of other tasks. } Furthermore, the results show a clear inconsistency in performance across different tasks. For the SOTA models, there is a distinct difficulty trend: performance is highest in Understanding, degrades in Generation, and slightly further in Editing, with a precipitous drop in the Interleaved Generation task, which demands the most complex reasoning. This suggests models are more adept at interpreting knowledge than generatively applying or manipulating it.
If a question cannot be correctly reasoned or understood by the MLLM component in UMMs, the visual generation or editing results cannot be correct either. Hence one can conclude that, the understanding capabilities of UMMs restrict the upper bound of corresponding visual generation and editing capabilities. These conclusion may inspire future research to design better MLLMs to make future UMMs handle complex or ambiguous instructions in practical applications. 

\subsection{Qualitative Results}
In addition to quantitative analysis, we also visualize the predictions from five state-of-the-art UMMs (\emph{i.e.},  Gemini Nano Banana~\cite{comanici2025gemini}, GPT-Image-1~\cite{hurst2024gpt4o}, Bagel-7B~\cite{deng2025bagel}, Ovis-U1~\cite{wang2025ovisu1}, and Qwen-Image~\cite{wu2025qwenimage}) on understanding, generation, and editing tasks for qualitative analysis. Such results are demonstrated in Fig.~\ref{fig:visualization_uge}. Compared to other open-source models, Gemini Nano Banana and GPT-Image-1 have better visual quality and image-text consistency, especially on generation and editing tasks. Moreover, Nano Banana has better main object consistency and text generation quality than GPT-Image-1, which shows superior generation and editing capabilities. These results are consistent with the quantitative results in Table~\ref{tab:model_comparison}. More visualization results will be shown in the suppl. 

\subsection{Empirical Analysis}

\begin{table}[t]
  \centering
      \caption{Human verification consistency of DCE, where U/G/E/I indicate understanding, generation, editing, and interleaved generation. The high consistency rate verifies the reliability of DCE.}
      \vspace{-1em}
      \label{table:gemini_human_consistency}
      \small
      \setlength{\tabcolsep}{5pt} 
     \renewcommand{\arraystretch}{4.0}
   { \fontsize{8.3}{3}\selectfont{
      \begin{tabular}{c|ccccc}
      \bottomrule
      Consistency (\%)&U&G&E&I&Overall\\
        \hline
         Gemini-2.5-Pro vs Human& 90.2 & 92.5 & 91.7 & 83.9 & 90.7 \\
        GPT-5  vs Human& 90.3 & 85.0 & 54.8 & 74.2 & 73.7 \\
      \toprule
      \end{tabular}
      }}
      \vspace{-2em}
  \end{table}

\noindent\textbf{Human verification of DCE. }To measure the reliability of our proposed DCE, we sample 10\% questions in {\dataset}, and manually verify the evaluation results by multiple experts. Then we calculate the percentage of judgement questions with the same decision as human verification consistency. The evaluation results are shown in Table~\ref{table:gemini_human_consistency} (upper), the overall consistency achieves 90.7\%, and the consistency in three majority tasks (\emph{i.e.}, understanding, generation, editing) is also higher than 90\%. 
These promising results illustrate the reliability of DCE.
Note that the consistency in the interleaved generation task is relatively lower. By manually analyzing these questions, we find that the complex ``image-text consistency check'' judgement questions in this task improves the evaluation difficulty. This finding motivates us for future research direction. 

\noindent\textbf{Choice of Judge Model. }Additionally, we are also curious about the choice of judge model used in DCE. Specifically, we use two widely used state-of-the-art MLLMs, \emph{i.e.}, Gemini-2.5-Pro~\cite{comanici2025gemini} and GPT-5~\cite{hurst2024gpt4o}, to evaluate the sampled responses with the same judgement questions, then calculate the human consistency rate respectively. The evaluation results are shown in Table~\ref{table:gemini_human_consistency}. According to the results, though GPT-5 has the same or similar judgement consistency on understanding (90.3 vs. 90.2) and generation (85.0 vs. 92.5) tasks, its consistency on judging editing and interleaved generation tasks is still far behind that of Gemini-2.5-Pro. A feasible explanation is that Gemini-2.5-Pro has better reasoning capabilities, which indicates that it can more precisely capture the visual details and differences between predictions and references. Therefore, the consistency of Gemini-2.5-Pro on editing and interleaved generation are much higher than that of GPT-5. 
These results verify our choice in DCE design, and also inspire furture research direction in complex visual relationship analysis. 

\noindent\textbf{Does state-of-the-art UMMs obtains ``world knowledge''? }After verifying the validity of DCE, one can further analyze whether these excellent models possess world knowledge. As shown in Table~\ref{table:clear_prompt}, we conducted an ablation study by replacing ambiguous, reasoning-intensive prompts with clear prompts across four tasks: understanding (U), generation (G), editing (E), and interleaved (I). This adjustment removed the need for the models to perform the most challenging reasoning steps, yet the results in Table 5 show consistent performance improvements across all tasks. This enhancement can be attributed to the activation of the models' inherent world knowledge, which was better utilized when clear prompts were provided.
Further analysis of module-specific issues will be detailed in the supplementary materials. 

\noindent\textbf{Analysis of different reasoning types. }
As shown in Table~\ref{table:reasoning_type_analysis} that different types of reasoning can also lead to variations in the difficulty of the problem. The ablation results highlight the varying difficulty of reasoning tasks requiring world knowledge across spatial, temporal, causal, comparative, analogical, logical, and no reasoning types. Gemini Nano Banana consistently achieves the best performance, followed by GPT-Image-1, while Bagel w/ CoT struggles significantly across all categories. Temporal and causal reasoning emerge as the most challenging tasks, reflecting the complexity of encoding sequences and relationships, whereas tasks requiring no reasoning are the easiest. These results emphasize the need for improved multimodal reasoning frameworks to address weaknesses in handling complex reasoning types.

\begin{table}[t]
  \centering
      \caption{Performance comparison of Gemini Nano Banana and GPT-4o with and without clear prompts across four tasks: understanding (U), generation (G), editing (E), and interleaved (I).}
      \vspace{-1em}
      \label{table:clear_prompt}
      \small
      \setlength{\tabcolsep}{5pt} 
     \renewcommand{\arraystretch}{4.0}
   { \fontsize{8.3}{3}\selectfont{
      \begin{tabular}{c|ccccc}
      \bottomrule
      Performance (\%)&U&G&E&I&Overall\\
        \hline
         Gemini Nano Banana & 61.7 & 45.9 & 53.7 & 45.1 & 52.9 \\
        + Clear prompt & 72.9 & 61.3 & 64.3 & 56.4  &  65.2 \\

        \hline
        GPT-4o+GPT-Image-1  & 50.2 & 44.2 & 45.9 & 35.5 & 45.7 \\
        + Clear Prompt & 63.2 & 61.7 & 57.7 & 52.4 & 60.0 \\
      \toprule
      \end{tabular}
      }}
      \vspace{-1em}
  \end{table}

\begin{table}[t]
  \centering
      \caption{Performance comparison of three UMMs across different reasoning types, \emph{i.e.}, spatial, temporal, causal, comparative, analogical, logical, and no reasoning tasks.}
      \vspace{-1em}
      \label{table:reasoning_type_analysis}
      \small
      \setlength{\tabcolsep}{2pt} 
     \renewcommand{\arraystretch}{3.0}
   { \fontsize{8.3}{3}\selectfont{
      \begin{tabular}{c|ccc}
      \bottomrule
      Reasoning Type& Gemini Nano Banana & GPT-Image-1&Bagel w/ CoT \\
        \hline
         Spatial & 51.2 & 43.9&  25.4 \\
        Temporal & 57.3 & 53.0& 25.9 \\

        Casual & 56.2 & 47.1& 25.5 \\
        Comparative & 60.9 & 47.5& 24.2 \\
        Analogical & 46.4 & 43.6& 23.9 \\
        Logical & 49.9 & 42.5& 18.3 \\
        No Reasoning & 52.8 & 50.9& 21.6 \\
      \toprule
      \end{tabular}
      }}
      \vspace{-2em}
  \end{table}
\section{Conclusion}
We present {\dataset}, a comprehensive benchmark to assess the visual understanding, generation, editing, and interleaved generation capabilities of unified multimodal models (UMMs) and generative models across a broad scope of world knowledge and reasoning types. By a human-in-the-loop construction and annotation strategy, {\dataset} shows the merits of inter-task evaluation, reasoning-type effect investigation, and real-world application assessment. To ensure the fine-grained and concrete judgement, we propose a deterministic checklist-based evaluation, which leverages a series of atomic ``Y/N'' judgement questions to assess a deterministic key part of answers, thereby improving the reliability of the LLM-as-a-Judge framework. Extensive experiments reveal that most UMMs exhibit severe world knowledge deficits and struggle significantly with complex reasoning. However, we also find that these deficits can be partially mitigated by simple plug-in reasoning modules, offering a promising direction for developing more robust future models.
{
    \small
    \bibliographystyle{ieeenat_fullname}
    \bibliography{main}
}
\clearpage
\appendix
In Sec.~\ref{sec:topic_category}, we illustrate the details of covered topic type in AEGIS. In Sec.~\ref{sec:reason_category}, we explain the details of each covered reasoning type in AEGIS. In Sec.~\ref{sec:more_ablation}, we conduct more in-depth analysis to reveal the key results in world knowledge evaluation. And in Sec.~\ref{sec:templates}, we illustrate the essential prompts used in AEGIS. 

\section{AEGIS Topic Type Descriptions}
\label{sec:topic_category}

AEGIS organizes real‑world knowledge into three domains (STEM, Humanities, Daily Life) and subdivides each topic into finer sub-topics to assess complementary facets.

\subsection{STEM}

The STEM topic assesses proficiency in Science, Technology, Engineering, and Mathematics, focusing on quantitative reasoning, application of physical and mathematical principles, and problem solving grounded in formal methods. It includes:

\begin{itemize}
    \item \textbf{Biology} assesses knowledge related to biological common sense, including representative species and ecological traits, fundamental life processes, and biological concepts that carry cultural relevance. \textit{Example: Please draw a picture of a female modern relative in Asia of the animal in \texttt{<image1>} .}
    \item \textbf{Chemistry} focuses on chemical substances, everyday chemical phenomena, and chemistry embedded in traditional crafts, covering common substances’ uses, safety awareness, and widely known processes across cultures and industries. \textit{Example: Given the two compounds shown in \texttt{<image1>} and \texttt{<image2>}, what is the expected reaction product in concentrated sulfuric acid?}
    \item \textbf{Mathematics} evaluates understanding of foundational mathematical concepts, common geometric figures, and everyday applications, emphasizing arithmetic, measurement, and shape recognition commonly taught across cultures. \textit{Example: The image \texttt{<image1>} shown represents a cubic function. If the coefficient of the cubic term is 1/2, what is the coefficient of the linear term?}
    \item \textbf{Medicine} examines basic medical and health literacy, including disease prevention, first aid fundamentals, and concepts or tools widely recognized in both traditional and modern healthcare practices. \textit{Example: How does salicylic acid \texttt{<image1>} enhance therapeutic efficacy? please draw the Chemical bond-line formula of the improved drug.}
    \item \textbf{Physics} focuses on everyday physical phenomena and foundational concepts—mechanics, thermodynamics, electromagnetism, optics, and acoustics—highlighting intuitive, real-world applications and explanations. \textit{Example: According to the principle of thin-film interference, please color the blank areas in the diagram \texttt{<image1>}.}
    \item \textbf{Astronomy \& Geography} assesses recognition of typical celestial and geographic features, including naked-eye sky phenomena, seasonal and directional knowledge, and culturally emblematic landmarks and biomes. \textit{Example: \texttt{<image1>} shows what a location at 60 degrees north latitude looked like before 1908. Please draw what the same location looked like after 1908.}
    \item \textbf{IT} focuses on common digital literacy and information technology concepts, including basic computing and networking, routine data and security practices, and widely used software/hardware terms. \textit{Example: \texttt{<image1>} shows a diagram of the CPU architecture. Please use the same color scheme to draw a diagram of the architecture of another common computing chip.}
\end{itemize}

\subsection{Humanities}

The Humanities topic evaluates understanding of human society, culture, and creative expression, emphasizing interpretive reasoning, historical analysis, contextual understanding, and critical evaluation of artifacts and practices. It includes: 

\begin{itemize}
    \item \textbf{Agriculture} evaluates knowledge of agricultural practices, crops, tools, and food systems across regions, including traditional and modern methods and their cultural-economic significance. \textit{Example: Using the label provided in the image \texttt{<image1>}, please colour the map \texttt{<image2>} according to the proportion of hybrid rice cultivated relative to total rice acreage.}
    \item \textbf{History} examines recognition of major historical events, periods, figures, and artifacts, emphasizing chronology, causation, and cultural impact. \textit{Example: Which event in the Qing Dynasty is similar to the one shown in this image \texttt{<image1>}?}
    \item \textbf{Movie} assesses familiarity with influential films, genres, directors, iconic scenes, and culturally significant cinematic symbols. \textit{Example: What is the MacGuffin of the 1942 Academy Award for Best Original Screenplay?}
    \item \textbf{Music} focuses on musical traditions, instruments, genres, and notable composers or performers, highlighting stylistic features and cultural contexts. \textit{Example: Generate a simple sheet music score of 'Twinkle Twinkle Little Star' in C major.}
    \item \textbf{Art} evaluates understanding of visual arts, styles, techniques, movements, and canonical works or artists across cultures and eras. \textit{Example: Edit \texttt{<image1>} to show Mona Lisa looking away with a disdainful expression and holding up a sign indicating she doesn't want her photo taken.}
    \item \textbf{Culture} assesses broader cultural practices, norms, heritage items, and symbols that define collective identities and social life. \textit{Example: Replace the outer skin of the three pastries in the middle of \texttt{<image1>} with the style of North China.}
    \item \textbf{Architecture} tests recognition of architectural styles, structural features, landmark buildings, and the historical-technological contexts of the built environment. \textit{Example: What is another religious sightseeing location in the same city as \texttt{<image1>}?}
\end{itemize}

\subsection{Daily Life}

The Daily fife topic evaluates practical knowledge and daily reasoning in common modern contexts, emphasizing situational understanding, routine decision making ‐ and the recognition of tools, activities and media encountered in daily environments.

\begin{itemize}
    \item \textbf{Activity} evaluates familiarity with common daily activities and leisure practices, including their typical tools, settings, and procedural steps. \textit{Example: Based on this screenshot \texttt{<image1>}, is there any Argentine player in an offside position?}
    \item \textbf{Anime} assesses recognition of notable anime series, characters, visual tropes, and stylistic conventions, as well as culturally salient symbols in animated media. \textit{Example: Where did the protagonist of One Piece go after bidding farewell to the Empress and before witnessing his brother's death?}
    \item \textbf{Game} focuses on understanding of video and tabletop games, including iconic titles, genres, gameplay elements, and distinctive in‑game artifacts or interfaces. \textit{Example: What is another well-known game produced by the team leader of the 2022 TGA Game of the Year for Mobile?}
    \item \textbf{Photography} examines knowledge of photographic equipment, techniques, genres, and visual conventions used in image capture and editing workflows. \textit{Example: Draw the photo of the girl \texttt{<image1>}, from the illustration to the cosplay photo from the most famous anime expo in the world. But keep the original background.}
    \item \textbf{Engineering} evaluates practical understanding of daily engineering artifacts, mechanisms, household devices, and basic technical operations relevant to daily environments. \textit{Example: Add appropriate materials to \texttt{<image1>} to make it a simple distiller.}
    \item \textbf{Food} assesses recognition of ingredients, dishes, cooking methods, dining customs, and nutrition concepts commonly encountered in daily meals. \textit{Example: Add Sichuan characteristics to this dish \texttt{<image1>}.}
    \item \textbf{Traffic} tests the ability to identify transportation modes, road signs, traffic rules, and navigation conventions used in urban mobility. \textit{Example: Draw the fastest rail transit route from the tower location to the red dot location in \texttt{<image1>}.}
\end{itemize}

\section{AEGIS Reasoning Type Descriptions}
\label{sec:reason_category}
Beyond general world knowledge, AEGIS further probes LLM’s capacity to follow obfuscated instructions by evaluating its underlying reasoning skills. Specifically, AEGIS categorizes reasoning into six types: 

\begin{itemize}
    \item \textbf{Spatial Reasoning} evaluates the ability to infer relationships involving position, distance, orientation, containment, and part–whole layout in 2D/3D space, which accounts for 10.9\% of the entire benchmark. \textit{Example: Given the front view \texttt{<image1>}, top view \texttt{<image2>}, and right side view \texttt{<image3>} of a 3D object, draw a picture of its isometric projection.}
    \item \textbf{Temporal Reasoning} assesses understanding of temporal order, duration, concurrency, and schedules, including before/after relations and timeline consistency, which accounts for 12.2\% of the entire benchmark. \textit{Example: Edit it to show how \texttt{<image1>} looks today.}
    \item \textbf{Causal Reasoning} examines the ability to identify cause–and–effect relations, necessary/sufficient conditions, and outcomes of interventions or counterfactual changes, which accounts for 12.2\% of the entire benchmark. \textit{Example: Infer a unified astronomical event based on \texttt{<image1>} and \texttt{<image2>}.}
    \item \textbf{Comparative Reasoning} concerns any comparison involving two or more entities along one or multiple dimensions, and drawing conclusions based on their relative differences or rankings, which accounts for 15.4\% of the entire benchmark. \textit{Example: \texttt{<image1>} and \texttt{<image2>}, which requires more cooking steps?}
    \item \textbf{Analogical Reasoning} evaluates mapping relational structure from a known scenario to a novel one, recognizing proportional or functional analogies, which accounts for 9.4\% of the entire benchmark. \textit{Example: Just as \texttt{<image1>} is to his corresponding anime work, who is the character in Naruto occupying a similar position?}
    \item \textbf{Logical Reasoning} emphasizes drawing conclusions that follow coherently from stated facts, rules, or constraints in everyday contexts, which accounts for 36.3\% of the entire benchmark. \textit{Example: Draw a Venn diagram with three intersecting sets A, B, and C, and shade the region corresponding to (A $\cap$ B) $\cup$ C.}
\end{itemize}

\begin{table*}[t]
\centering
\caption{
    Comprehensive Performance Comparison for Gemini Nano Banana (Gemini for short) and GPT-4o with GPT-Image-1 (GPT for short) with different types of prompts and external web search tools.
}
\label{tab:prompt_exp}
\setlength{\tabcolsep}{3pt} %
    \renewcommand{\arraystretch}{3}%
    { \fontsize{8.3}{3}\selectfont{
\begin{tabular}{l| ccc| ccc| ccc |ccc| @{\hspace{1.5em}} c}
\toprule
\multirow{2}{*}{\textbf{Model}} & \multicolumn{3}{c}{\textbf{Understanding}} & \multicolumn{3}{c}{\textbf{Generation}} & \multicolumn{3}{c}{\textbf{Editing}} & \multicolumn{3}{c}{\textbf{Interleaved Generation}} & \multirow{2}{*}{\textbf{Overall}} \\
\cmidrule(lr){2-4} \cmidrule(lr){5-7} \cmidrule(lr){8-10} \cmidrule(lr){11-13}
 & STEM & Humanity & Life & STEM & Humanity & Life & STEM & Humanity & Life & STEM & Humanity & Life & \\
\midrule

Gemini & 64.5 & 65.7 & 55.0 & 42.6 & 49.5 & 45.5 & 44.4 & 62.4 & 54.2 & 50.2 & 41.6 & 43.4 & 52.9 \\
Gemini w/ Web Search & 64.9 & 69.2 & 60.0 & - & - & - & - & - & - & - & - & - & - \\
Gemini w/ GPT prompt & 63.8 & 64.7 & 56.2 & 37.8 & 46.0 & 44.3 & 41.0 & 61.4 & 51.3 & 47.3 & 43.6 & 43.3 & 51.1 \\
Gemini w/ Gemini prompt & 63.4 & 72.9 & 57.6 & 45.3 & 57.9 & 52.2 & 42.2 & 63.9 & 47.8 & 50.8 & 49.6 & 47.2 & 55.2 \\
Gemini w/ Clear prompt & 72.4 & 74.3 & 72.1 & 53.6 & 67.6 & 62.6 & 54.2 & 68.4 & 70.4 & 53.1 & 62.0 & 54.0 & 65.2 \\
Gemini w/ Clear \&  Web & 72.8 & 80.8 & 73.3 & - & - & - & - & - & - & - & - & - & - \\
\midrule

GPT & 52.9 & 50.9 & 46.9 & 38.2 & 51.6 & 42.8 & 39.4 & 53.2 & 45.2 & 38.9 & 34.7 & 33.0 & 45.7 \\
GPT w/ GPT prompt & 48.4 & 52.0 & 43.4 & 35.6 & 52.8 & 41.9 & 38.1 & 54.9 & 46.0 & 33.5 & 33.6 & 34.7 & 44.7 \\
GPT w/ Gemini prompt & 57.0 & 66.7 & 53.2 & 42.6 & 57.3 & 47.0 & 39.0 & 59.3 & 45.4 & 44.6 & 36.1 & 41.8 & 50.8 \\
GPT w/ Clear prompt & 61.6 & 65.4 & 62.7 & 52.1 & 71.5 & 61.6 & 49.2 & 66.9 & 56.9 & 51.0 & 57.2 & 49.2 & 60.0 \\

\midrule

 Gemini-3-Pro & 77.7 & 79.3 & 70.4 & 62.2 & 64.4 & 58.2 & 64.1 & 67.8 & 58.5 & 42.6 & 40.2 & 38.5 & 64.3 \\

\bottomrule
\end{tabular}
}}
\end{table*}

\section{Additional Experiments}\label{sec:more_ablation}

In this section, we provide additional experiments based on Gemini Nano Banana~\cite{comanici2025gemini} (Gemini for short) and GPT-4o with GPT-Image-1~\cite{hurst2024gpt4o} (GPT for short) to examine how prompts of varying specificity affect performance across tasks. We further disentangle common-sense knowledge from the UMMs via controlled ablations to isolate module-specific issues and quantify their impact. We also investigate the upperbound of UMMs by evaluating the state-of-the-art Gemini-3-Pro (\emph{i.e.}, Nano Banana Pro).

\subsection{Evaluation on UMM Rewritten Prompts}

Beyond investigating the impact of external reasoning modules~\cite{cot} on Bagel~\cite{deng2025bagel}, we conducted an ablation study to further isolate the effect of reasoning quality. Specifically, we employed a ``self-reasoning'' strategy wherein the model first rewrites the raw prompt to generate a ``clear prompt,'' thereby mitigating the need for downstream reasoning by resolving ambiguities, identifying entities, and making implicit context explicit. 
Surprisingly, while manually verified clear prompts generally yield substantial gains, we observed \textbf{divergent effects} with self-rewriting: prompts rewritten by GPT-4o~\cite{hurst2024gpt4o} resulted in performance degradation compared to raw inputs, whereas those rewritten by Gemini Nano Banana~\cite{comanici2025gemini} led to performance improvements. 
To validate this disparity, we performed a cross-model evaluation by swapping the rewritten prompts, \emph{i.e.}, feeding GPT-4o with Gemini-rewritten prompts and vice versa. The results were consistent: GPT-generated rewrites caused performance drops across models, while Gemini-generated rewrites consistently yielded gains. 
These findings strongly suggest that Gemini Nano Banana possesses superior reasoning capabilities for instruction disambiguation compared to GPT-4o. Consequently, leveraging LLMs with advanced reasoning capabilities offers a promising avenue to mitigate the challenges posed by ambiguous or reasoning-intensive instructions in UMMs, thereby benefiting diverse tasks across a broad spectrum of world knowledge~\cite{niu2025wise}.

\subsection{Gemini-3-Pro has better World Knowledge}
As discussed in Sec.~\ref{sec:exp}, introducing training data with better quality and more abundant domain aspects can improve the world knowledge understanding abilities. However, those of the other tasks are not verified. Therefore, we evaluate Gemini-3-Pro, \emph{i.e.}, the extended version of Gemini Nano Banana, on AEGIS benchmark. As shown in the bottom of Table~\ref{tab:prompt_exp}, even using the reasoning-enhanced prompts for inference, the overall performance of Gemini-3-Pro (64.3) is still much higher than that of Gemini (52.9), and achieves comparable performance with Gemini using clear prompts. These promising results indicate that better pretraining data benefits to the world knowledge capabilities, and also show that Gemini-3-Pro is the state-of-the-art UMM in world knowledge understanding and generation aspects. 

\subsection{Evaluation on Web Search}

Following our investigation into disambiguating complex instructions, we further explored methods to mitigate prediction errors and hallucinations by incorporating new external knowledge. Intuitively, integrating a search engine should provide the essential, up-to-date information required for accurate responses.
To assess this, we conducted an ablation study evaluating the impact of web search augmentation on model performance. Specifically, we enabled the Google Search tool for Gemini (specifically, Gemini-2.5-Flash-Image) to ground its responses in current events and verifiable web-based facts. 
Counterintuitively, activating web search only results in marginal performance improvement in understanding tasks. Especially, for questions in STEM topics, the performance gain has only 0.4. A plausible explanation is that while search tools effectively acquire external world knowledge—crucial for verifying facts or retrieving recent events—they do not inherently strengthen the model's core reasoning capabilities. This finding aligns with human problem-solving behaviors: effective solutions rarely emerge from directly querying a complex, raw problem into a search engine. Instead, successful problem solving typically necessitates an initial reasoning phase to formulate clear, targeted queries before consulting external resources.
To verify our hypothesis, we further integrate the web search tool into Gemini with clear prompts, to investigate whether clearer and more effective problem description leads to more precise search results as auxiliary knowledge. As shown in Table~\ref{tab:prompt_exp}, by easing the problem with clear prompts, the understanding performance laragely increases. Especially, the performance gain of humanity and life understanding questions are both larger than 10.0, which indicates web search tools can improve the world knowledge capabilities of UMMs with clear problem descriptions. These results also imply the significance of inherent reasoning capabilities of UMMs during inference. 

\subsection{Investigation into Module-Specific Bottlenecks}
\label{sec:module_investigation}

\begin{figure*}
\begin{center}
\includegraphics[width=0.99\textwidth]{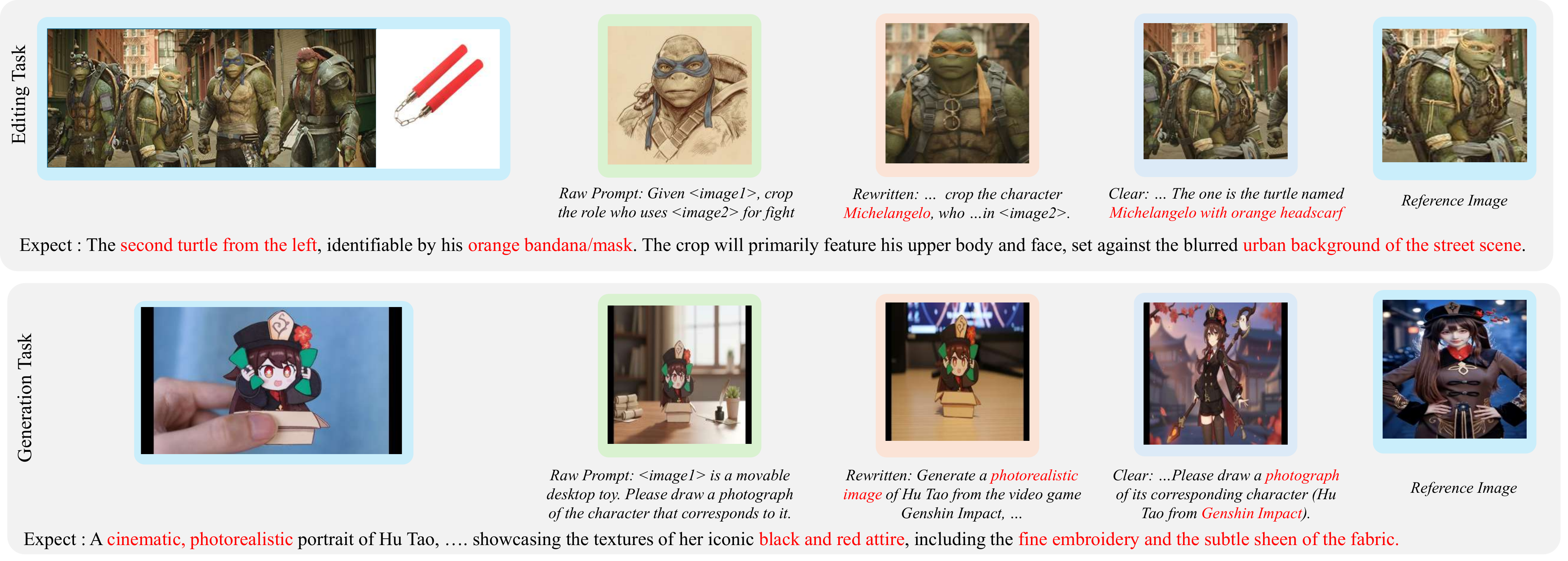}
\end{center}
\caption{Visualization of failure cases with raw and LLM rewritten prompts. We highlight the keypoints in the answers by \textcolor{red}{red color}. 
Though external reasoning modules (\emph{e.g.}, Gemini) can ease the generation difficulty by rewritting complex prompts, there still exist gaps towards precise reasoning capabilities under diverse tasks across world knowledge aspects. 
}
\label{fig:visualization_raw_failure}
\end{figure*}

Furthermore, we aimed to identify which component acts as the \emph{primary limiting factor} for world knowledge capabilities in UMMs: the LLM component or the visual decoder component. 
To locate the source of errors, we analyzed failure cases from Gemini. Specifically, we utilized Gemini-2.5-Pro to rewrite the original prompts via a self-reflection procedure, ensuring all implicit knowledge was made explicit. We then fed these rewritten prompts back into Gemini for generation.
Fig.~\ref{fig:visualization_raw_failure} presents a comparison of images generated from raw prompts, rewritten prompts, and verified clear prompts. Crucially, we observed that the LLM component successfully articulated the key visual attributes in the rewritten text (\emph{e.g.}, correctly identifying "Hu Tao" or "Michelangelo"). However, the visual decoder failed to render these concepts consistently, deviating from both the clear-prompt outputs and the ground truth. 
This discrepancy suggests that the \emph{visual decoder} restricts world knowledge capabilities in UMMs, likely due to insufficient knowledge encoding within the decoder itself or extreme sensitivity to input phrasing.

\begin{figure*}
\begin{center}
\includegraphics[width=0.99\textwidth]{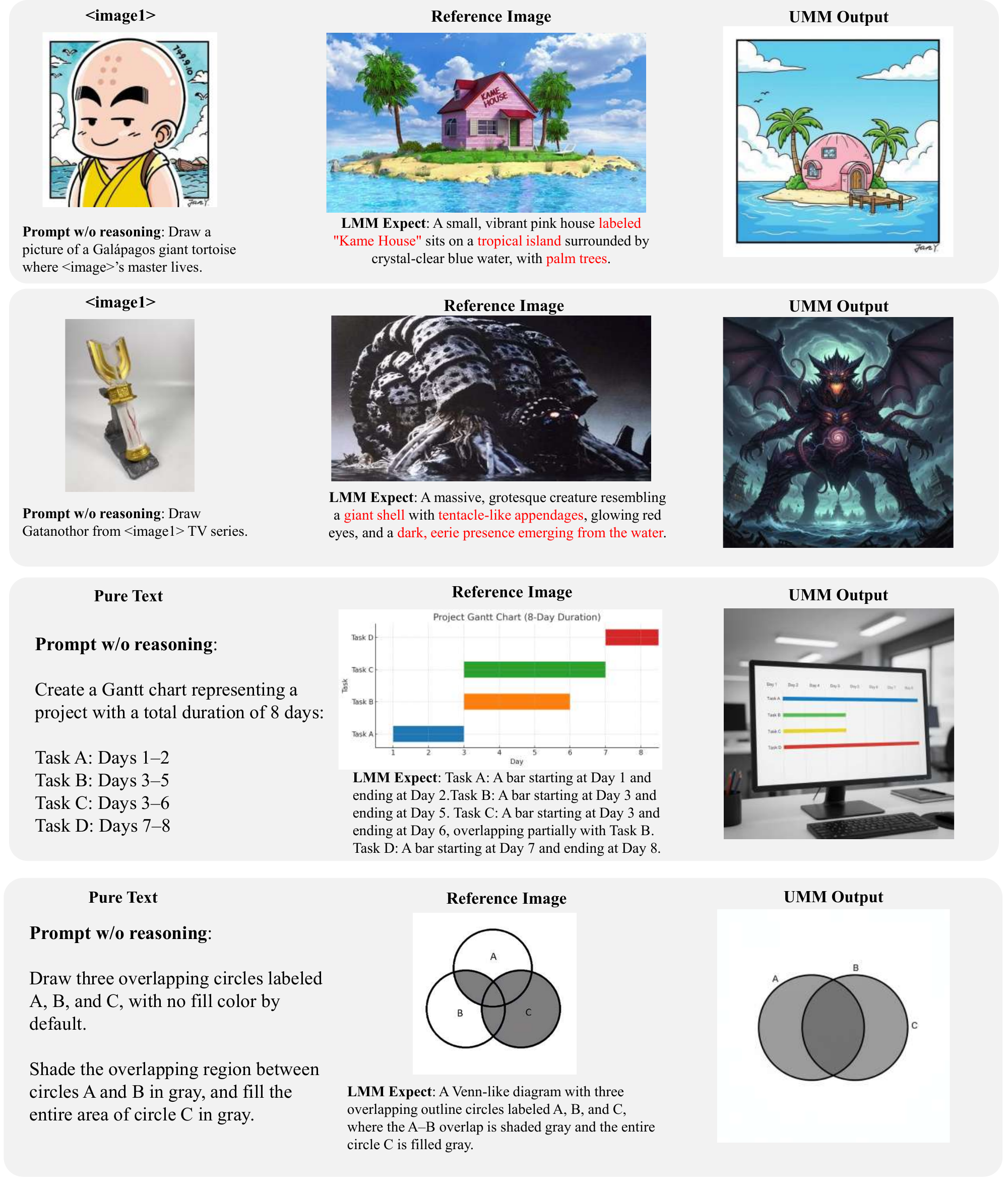}
\end{center}
\caption{
Visualization of failure cases with refined LLM descriptions of clear prompts from Gemini. One can find that even the descriptions precisely illustrate the answers, the visual decoder still usually struggles with generating correct answers. 
}
\label{fig:clear_prompt_failure_case}
\end{figure*}

To rigorously verify this hypothesis, we conduct a follow-up experiment using {extremely detailed descriptions}. We use Gemini-2.5-Pro to generate comprehensive visual descriptions that explicitly outline every keypoint required for generation or editing, effectively bypassing the model's need to recall visual attributes. We then feed these descriptions into Gemini. 
As shown in Fig.~\ref{fig:clear_prompt_failure_case}, despite the LLM providing highly accurate and detailed visual instructions, \emph{only the first} example shows a plausible result (with marginal shape discrepancies), while the others remained incorrect. These results definitively verify that the visual generation module is the bottleneck, identifying a misalignment between the model's strong textual understanding and its weaker visual generation capabilities, consistent with the performance gaps observed in \dataset.

\begin{figure*}
\begin{center}
\includegraphics[width=0.99\textwidth]{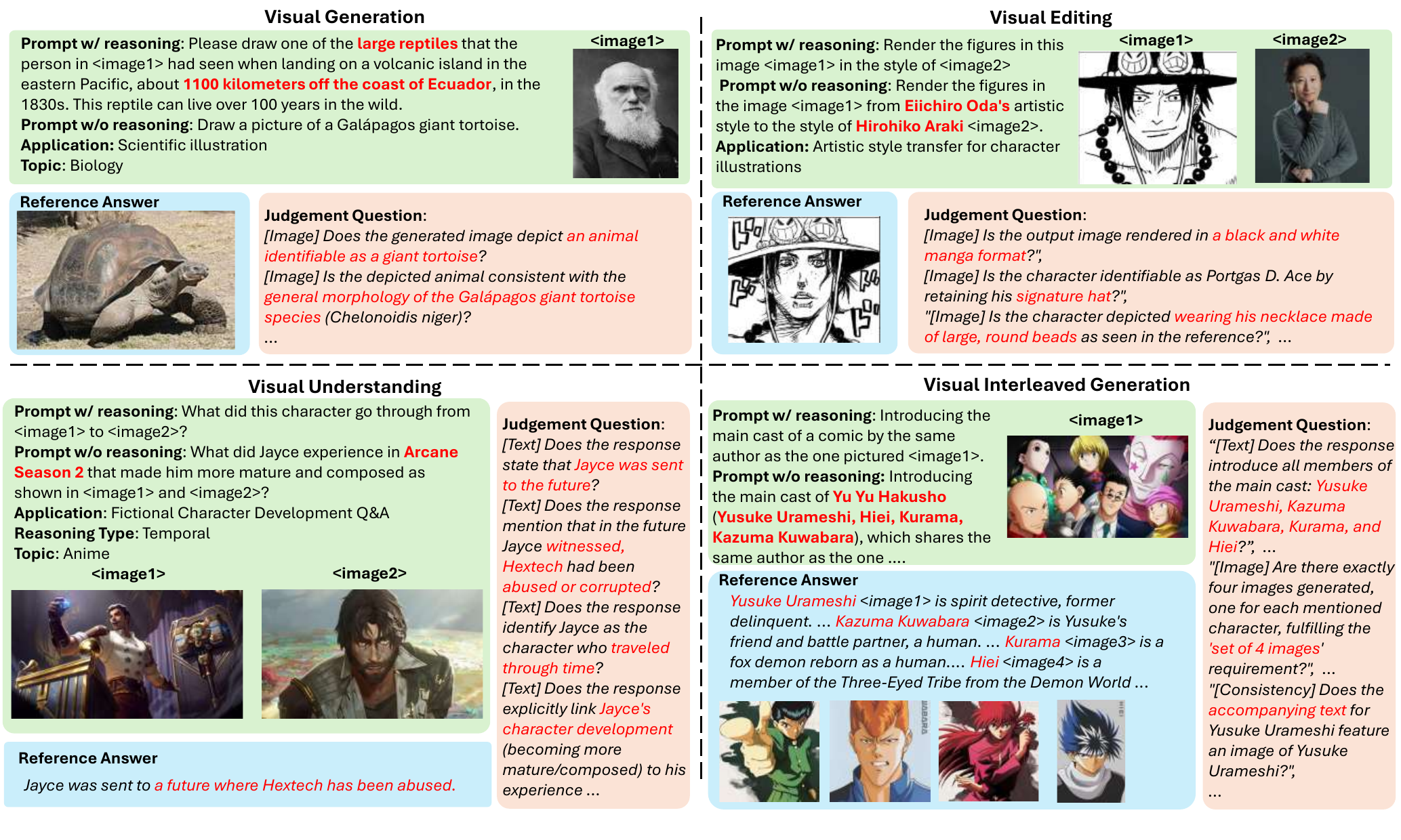}
\end{center}
\caption{Visualization of more questions in AEGIS benchmark. 
}
\label{fig:visualization_more}
\end{figure*}

\begin{figure*}
\begin{center}
\includegraphics[width=0.99\textwidth]{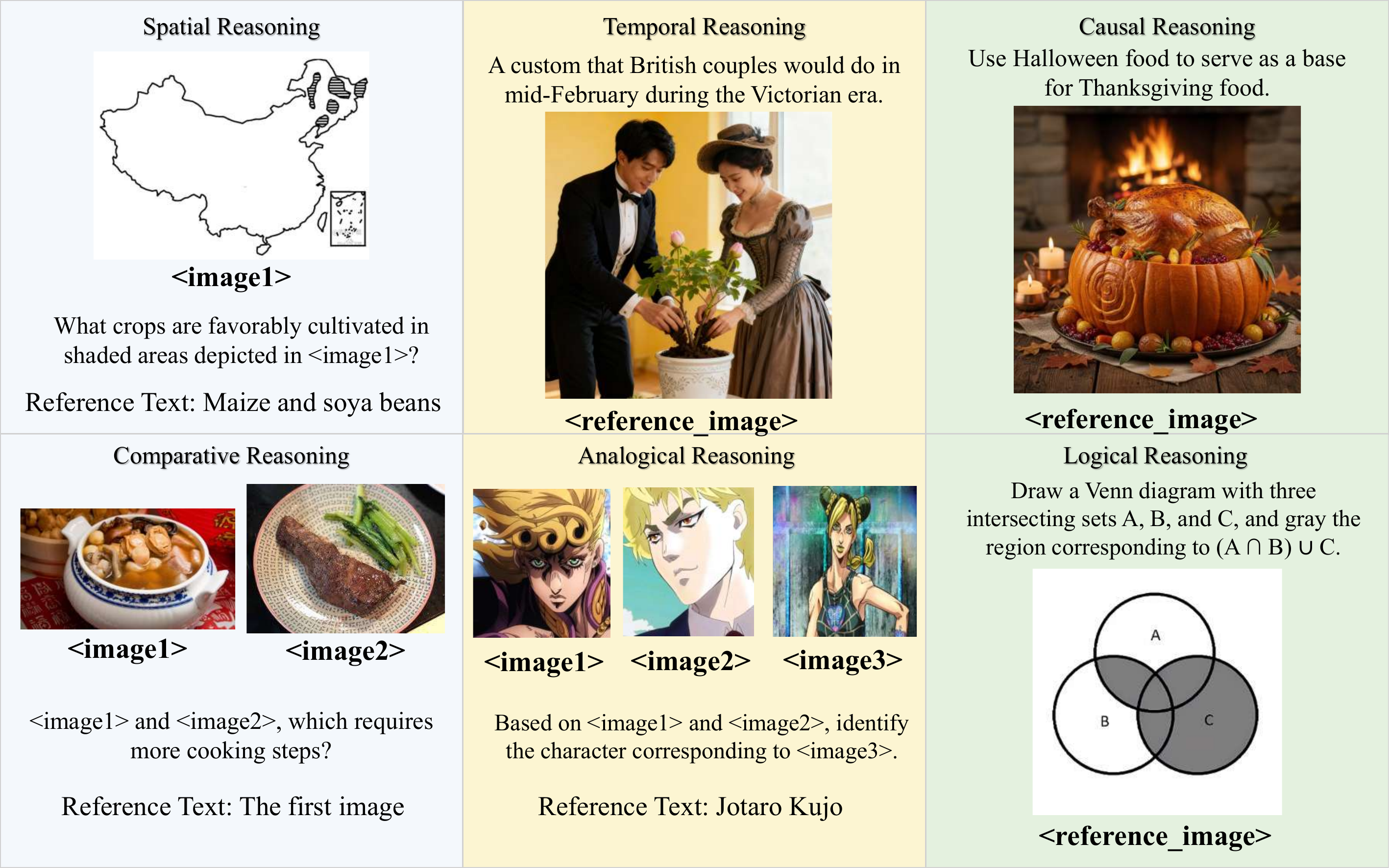}
\end{center}
\caption{
Visualization of questions with different reasoning types. AEGIS includes various reasoning types in questions, covering common scenarios in practical applications. 
}
\label{fig:visualization_reasoning_types}
\end{figure*}

\begin{figure*}
\begin{center}
\includegraphics[width=0.99\textwidth]{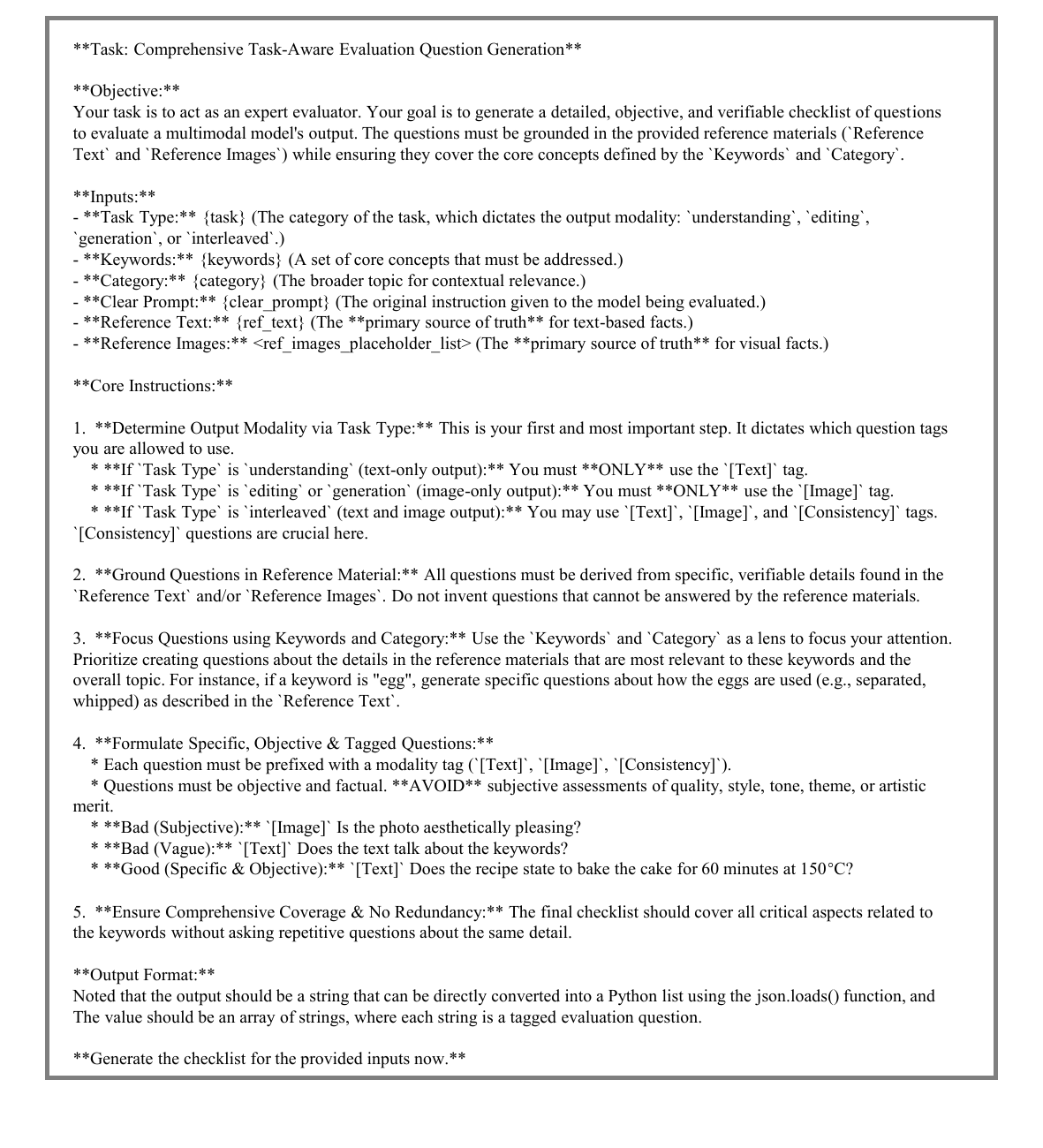}
\end{center}
\caption{Checklist generation prompts in AEGIS benchmark. We formulate the LLM-as-a-Judge evaluation by a series of atomic ``Y/N'' questions to avoid ambiguous judgments.
}
\label{fig:checklist_generation_prompt}
\end{figure*}

\begin{figure*}
\begin{center}
\includegraphics[width=0.99\textwidth]{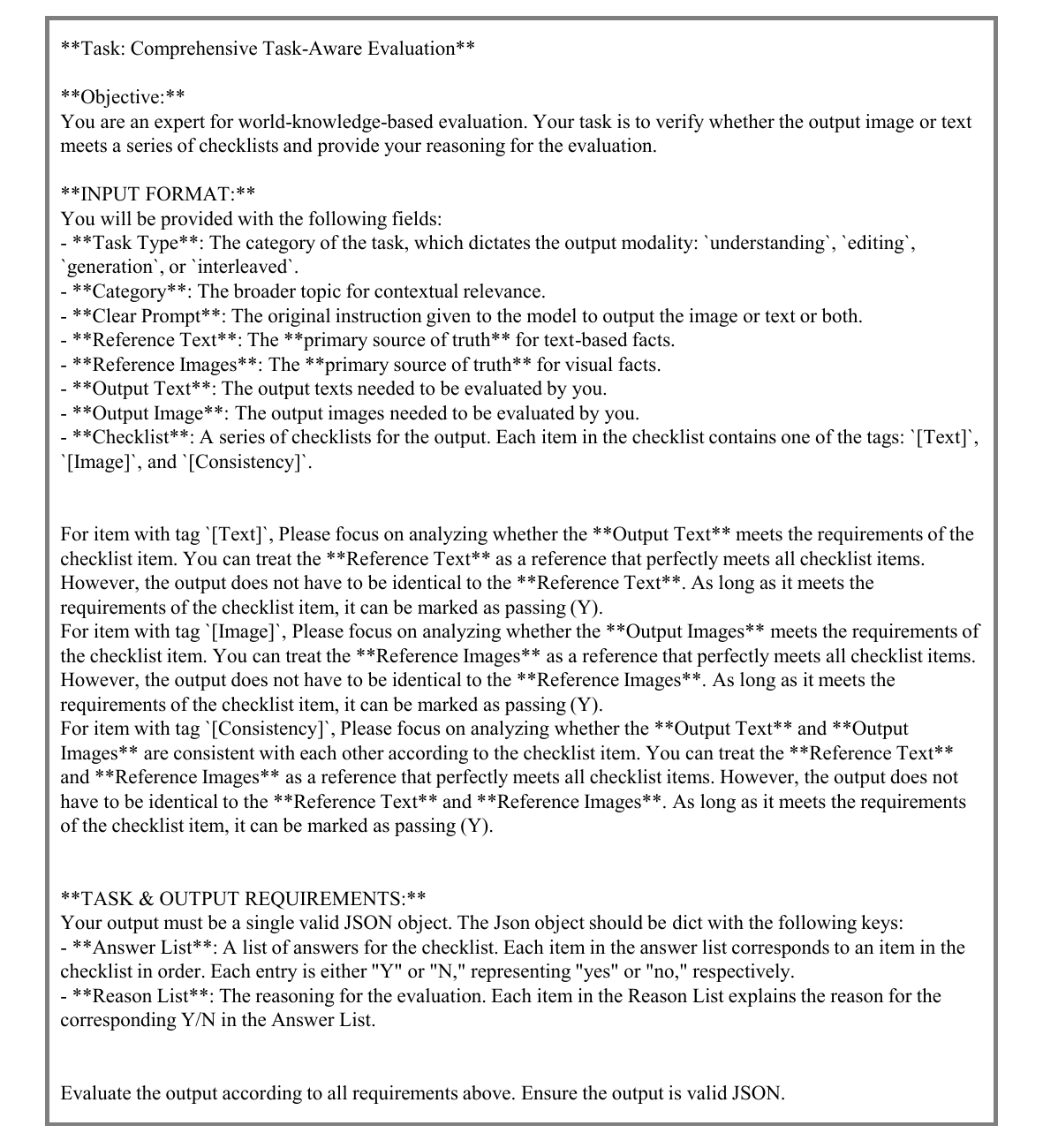}
\end{center}
\caption{DCE Evaluation prompts in AEGIS benchmark. We predict ``yes / no'' judgements for all atomic judgement questions, and calculate the percentage of ``yes'' judgements as final scores. 
}
\label{fig:evaluation_prompt}
\end{figure*}

\section{Essential Templates Used in AEGIS}\label{sec:templates}
Finally, we provide templates used in the AEGIS dataset annotations, including the checklist generation prompt in Fig.~\ref{fig:checklist_generation_prompt} and evaluation prompts in Fig.~\ref{fig:evaluation_prompt}. 

\end{document}